\documentclass[a4paper, 10pt, conference]{ieeeconf}      %

\IEEEoverridecommandlockouts                              %

\usepackage{amsmath}
\usepackage{amssymb}  %

\usepackage{bm}  %
\usepackage{graphicx}  %
\usepackage{subcaption} %
\usepackage{siunitx}
\usepackage{algpseudocode} %
\usepackage{algorithm} %
\usepackage{booktabs}  %
\usepackage{tabularx}  %

\usepackage{flushend} %
\usepackage[T1]{fontenc} %
\usepackage[utf8]{inputenc}
\usepackage{wrapfig} %
\usepackage[font=small]{caption} %

\usepackage[
        backend=biber,
        style=numeric,
        sorting=none,
        doi=false,
        isbn=false,
        url=false]{biblatex}

\newcommand{\dbm}[1]{\dot{\bm{#1}}}
\newcommand{\ddbm}[1]{\ddot{\bm{#1}}}

\newcommand{\eq}[1]{Eq.~(\ref{#1})}
\newcommand{\fig}[1]{Fig.~\ref{#1}}
\newcommand{\sect}[1]{Section~\ref{#1}}

\newcommand{\tabl}[1]{Table~\ref{#1}}
\newcommand{\alg}[1]{Algorithm~\ref{#1}}

\title{\LARGE \bf
Motion Macro Programming on Assistive Robotic Manipulators: Three Skill Types for Everyday Tasks
}

\author{Stefan Scherzinger$^{1}$, Pascal Becker$^{1}$, Arne Roennau$^{1}$ and R\"udiger Dillmann$^{1}$%
\thanks{
This research has been funded by the Federal Ministry of Education and Research of
Germany (BMBF) within the project ArNe in the framework of Robotic Systems in Health Care (project number 16SV8412)
\smallskip
\newline
 ${}^{1}$ All authors are with FZI Research Center for Information Technology, Haid-und-Neu-Str. 10-14, 76131 Karlsruhe, Germany
        {\tt\small \{scherzinger, pbecker, roennau, dillmann\}@fzi.de}}%
}

\begin{document}

\maketitle
\thispagestyle{empty}
\pagestyle{empty}

\begin{abstract}
Assistive robotic manipulators are becoming increasingly important for people with disabilities.
Teleoperating the manipulator in mundane tasks is part of their daily lives.
Instead of steering the robot through all actions, applying
self-recorded motion macros could greatly facilitate repetitive tasks.
Dynamic Movement Primitives (DMP) are a powerful method for skill learning via teleoperation. %
For this use case, however, they need simple heuristics to specify
where to start, stop, and parameterize a skill without a background in computer
science and academic sensor setups for autonomous perception.
To achieve this goal, this paper provides the concept of \textit{local, global}, and
\textit{hybrid} skills that form a modular basis for composing
single-handed tasks of daily living.
These skills are specified implicitly and can easily be programmed
by users themselves, requiring only their basic robotic manipulator.
The paper contributes all details for robot-agnostic implementations.
Experiments validate the developed methods for exemplary tasks, such as scratching an itchy spot,
sorting objects on a desk, and feeding a piggy bank with coins.
The paper is accompanied by an open-source implementation at
https://github.com/fzi-forschungszentrum-informatik/ArNe

\end{abstract}

\section{INTRODUCTION}
Robotic manipulators are becoming increasingly important as assistive devices
in home settings for people with disabilities.
Despite the complexity of tasks of daily living - which often require two
hands to accomplish -
there are tasks that users can accomplish on their own with one
robotic arm,
such as organizing things on their desk, or scratching an itchy spot on the knee.
Research shows promising hybrids between direct control and assistive
autonomy for safe drinking~\cite{Lillo2021}, object manipulation
\cite{Stoyanov2018},\cite{Lee2018}, and grasping \cite{Jain2016}, for instance.
This autonomy, however, usually comes at the expense of additional sensor and
hardware setups, e.g. \cite{Arrichiello2017}, that must be provided and
integrated into existing, basic manipulator systems.
Losey et al.~\cite{Losey2020} show an approach for embedding high-dimensional robot
behaviors into simplified, latent actions that users can then teleoperate.
Although improving users' performance in a kitchen setup, the behaviors must
be demonstrated by caregivers first. 

For the user group that can and wants to stay in cognitive control of their robot,
a simple mechanism for recording repetitive motion patterns and replaying them might already be a big support.
Steering the robot as a mere \textit{tool} through teleoperation, users could
flexibly create and personalize their own skill repertoire through programming by demonstration.
The principal contribution of the paper is a practical integration of
\textit{Dynamic Movement Primitives}
(DMPs)~\cite{Ijspeert2002},\cite{Ijspeert2013} into the field of assistive robot arms
for self-determined recording and playing back of repetitive motion patterns.
We address the important challenge of formulating start and goal states
in an intuitive way for the DMPs during teleoperation and simplify
much of the DMPs' complexity associated with time scaling.
We accompany the paper with an implementation for the
ROS~framework~\cite{Quigley2009}, which is available open-source\footnotemark[2]
\footnotetext[2]{https://github.com/fzi-forschungszentrum-informatik/ArNe}.
The remaining paper is structured as follows:
We briefly describe the principle mechanisms behind DMPs
in \sect{sec:theoretical_background}
to illustrate our simplifications to the original approach.
\sect{sec:manipulator_control} then describes our control interface for the manipulators.
The core of the paper is \sect{sec:composing_and_using_skills} with the
presentation of three skill types and their implementation.
We evaluate and discuss them in \sect{sec:experiments_and_results}, and
provide an overall conclusion in \sect{sec:conclusions}.

\section{BACKGROUND AND RELATED WORK}
\label{sec:theoretical_background}
\subsection{Dynamic Movement Primitives (DMPs)}
The principle idea of DMPs is to describe a trajectory $\bm{x}(t)$ of a state vector
$\bm{x} \in \mathbb{R}^n$ of $n$ dimensions with a set of ordinary differential equations.
In each dimension,
its scalar variable $x$ is defined by an individual spring-damper system, e.g.~\cite{Ijspeert2013}
\begin{equation}
        \label{eq:dmp_transformation_system}
        \tau \ddot{x} = D (K (g - x) - \dot{x} ) + f ~.
\end{equation}
\eq{eq:dmp_transformation_system}) is usually referred to as \textit{transformation system}.
One important feature of DMPs is the generalization
to new goals $g$, referred to as \textit{attractor states}. The duration
of the execution can be adjusted via the time scale $\tau$.
Both stiffness $K$ and damping $D$ are constants that need hand-tuning to specific use cases.
Likewise, the arrangement of terms of \eq{eq:dmp_transformation_system} also varies with use cases and implementations.
Popular enhancements include e.g. collision
avoidance~\cite{Pastor2009},~\cite{Hoffmann2009} and the integration of haptic
feedback~\cite{Pastor2012Towards}.
Lauretti et al. show a more recent approach for hybrid joint/Cartesian DMPs for redundant manipulators
with the advantage of maintaining human-like motion during collision avoidance\cite{Lauretti2019}.
The non-linear \textit{forcing term} $f(s(t))$ is the core of the framework, and is usually parameterized with a
\textit{phase variable} $s$ across the course of the skill.
This term is responsible for effecting desired motion characteristics,
which are to be learned from human-recorded training data.
All $n$ transformation systems with the shape of \eq{eq:dmp_transformation_system} are synchronized through $s$ according to the following
\textit{canonical system}
\begin{equation}
        \label{eq:dmp_canonical_system}
         \tau \dot{s}(t) = - \alpha s(t)
\end{equation}
that describes the correlation between phase time $s$ and real-time $t$ as an exponential decay from $1 \rightarrow 0$.
The parameter $\alpha$ is an additional constant for fine-tuning.

The perturbation $f$ is usually modeled with a superposition of radial basis functions, e.g.
\begin{equation}
        \label{eq:dmp_forcing_term}
        f(s) = \frac{\sum_i \omega_i \psi_i (s) s}{\sum_i \psi_i (s)} \quad ,
\end{equation}
with $\psi_i(s) = \text{exp}( -h_i (s - c_i)^2 )$ representing Gaussians
with constant center $c_i$ and constant width $h_i$.
The higher the number of basis functions, and the more \textit{clever} they overlap, %
the closer are they able to capture motion characteristics with $f$ in
\eq{eq:dmp_transformation_system}.
The choice of more suitable kernel functions can partially mitigate the
exploding computational complexity and can allow for an a-priori estimation of
the reproduction accuracy~\cite{Papageorgiou2018}.

If combined with Reinforcement Learning (RL) on robots with many degrees of
freedom, for instance, a massive number of basis functions can become a
substantial performance bottleneck, since all parameters need to be learned and
need to converge in numerous cycles of own trial and error~\cite{Colome2018}.

Learning human skills supervised into DMPs means learning
adjustable weights $\omega_i$ from example motions.
This is achieved by recording a skill-defining trajectory $x(t)$ %
of arbitrary duration from human demonstration.
Solving \eq{eq:dmp_transformation_system} for $f$ leads to a set of discrete
samples by substituting the trajectory points into
\begin{equation}
        \label{eq:dmp_target_samples}
        \widehat{f}(s(t)) = \ddot{x} - D( K(g - x) - \dot{x}) \frac{1}{\tau}~.
\end{equation}
This has a certain resemblance to an inverse dynamics approach,
in which $\widehat{f}$ as reaction force is uniquely defined by the motion profile over time.
The weights of \eq{eq:dmp_forcing_term} are then adjusted
with the samples obtained from \eq{eq:dmp_target_samples}, e.g. by
minimizing $\sum_s (\widehat{f}(s) - f(s))^2$ with multivariate linear regression.

Finally, DMPs are deployed as follows:
First \eq{eq:dmp_canonical_system} and then each
of the $n$ transformation systems with \eq{eq:dmp_transformation_system} are numerically integrated with a
task-specific $\tau$ as a solution to an initial value problem.
The combined state $\bm{x}(t)$ is then passed to a suitable robot controller as a reference signal for motion tracking.

\subsection{Assumptions and Simplifications}
In this paper we describe an approach for using
simplified DMPs as a tool to teleoperate assistive robot manipulators for repetitive tasks.
Our primary motivation is to provide the methods needed to implement a simple add-on
for assistive devices that users might deploy at home. The open-source implementation
in the ROS framework should be considered an inspiration, and we hope that it
can be used to implement custom versions in different programming languages.
From a technical perspective, it is worth pointing out that the core feature of generalization between different \textit{attractor states} is
achieved by the dynamical system itself, and solely depends on the shape of
\eq{eq:dmp_transformation_system}.
In contrast, the principle motivation for parameterizing $f$ with overlapping basis functions is to
provide a mechanism for time-independent scaling.
An upper limit of accuracy would be an infinite number of basis functions.
There is a trade-off between capturing details in the motion profiles and being computationally tractable.
Also note that it's very difficult to distinguish between noise in $f$ and
characteristics that yield fine-grained detail on position level after numerical
integration. The accuracy of the recording thus inherently suffers from approximating basis functions.
Our idea is as follows: If we use the same
number of samples for describing recorded motion with
\eq{eq:dmp_target_samples} as for numerically integrating
\eq{eq:dmp_transformation_system},
then we can omit the parameterization and hence accuracy-limiting behavior.
Temporal scaling would then be achieved with a mere assignment of timestamps to
the computed motion. 
For instance, if we recorded a motion of \SI{10}{s} at \SI{100}{Hz} ($= 1000$ samples) and
wanted to execute that in a different configuration during \SI{20}{s}, we would
numerically integrate \eq{eq:dmp_transformation_system} with
$1000$ steps and follow these setpoints with an interpolating controller at half the rate at \SI{50}{Hz}.

Our simplifications to the DMP approach summarize as follows:
\begin{itemize}
\item Set $\tau \equiv 1$ in \eq{eq:dmp_transformation_system}
\item Drop  \eq{eq:dmp_canonical_system}, \eq{eq:dmp_forcing_term}, and the phase variable $s$
\item Use $\widehat{f}$ from \eq{eq:dmp_target_samples} directly to integrate \eq{eq:dmp_transformation_system}
\end{itemize}

The last point can mean a substantial boost to computational performance: In comparison to linear
regression on a possibly huge amount of samples, \eq{eq:dmp_target_samples} is
an analytic expression that is directly computed from the recording.
In the remainder of the paper, we apply these simplifications to our use case of teleoperating assistive robotic
manipulators with motion skills.

\section{MANIPULATOR CONTROL}
\label{sec:manipulator_control}

This section describes our methods for teleoperation and motion tracking.
We target 6-axis light-weight robotic manipulators that are common as assistive devices.
All grippers are suitable as long as their state can be mapped to a single degree of freedom.
\fig{fig:robot_coordinates} illustrates one such manipulator with three important coordinate frames:
\begin{figure}
        \centering
        \includegraphics[width=0.25\textwidth]{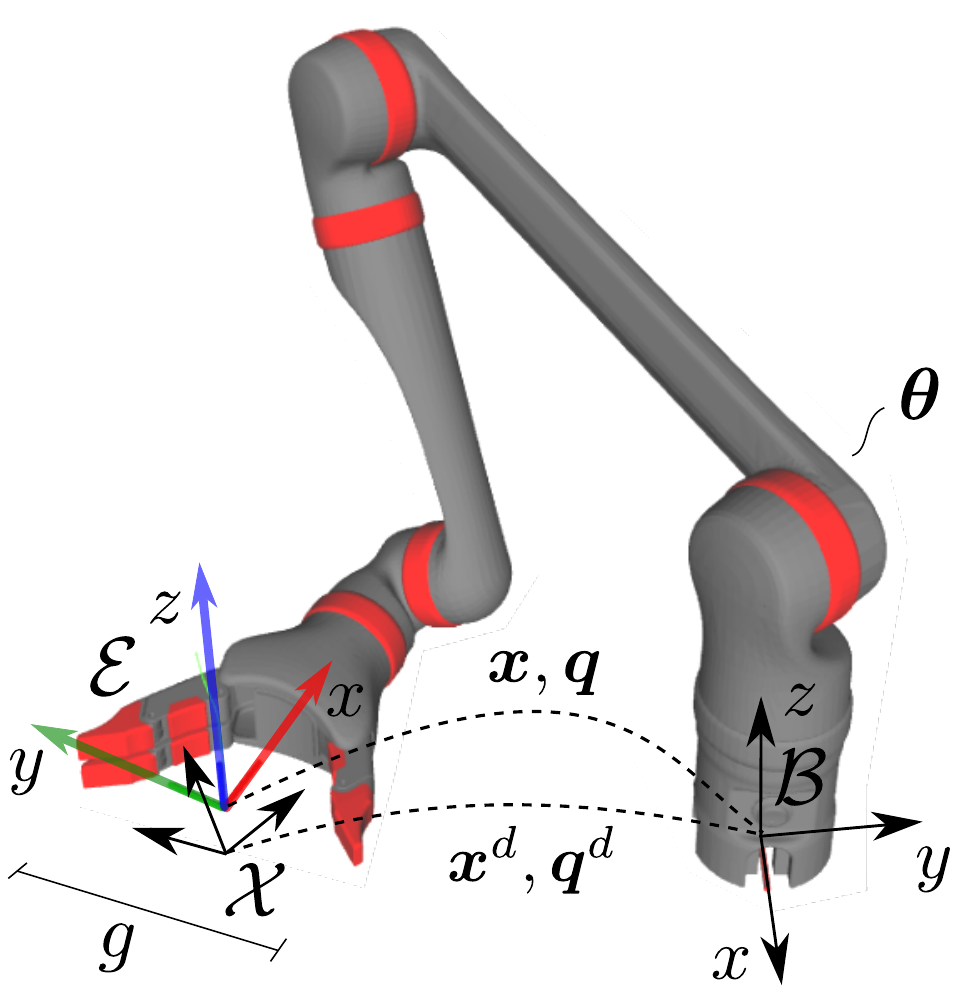}
        \caption{Schematic illustration of an exemplary 6-axis light-weight robotic
        manipulator with base frame $\mathcal{B}$, end-effector frame
        $\mathcal{E}$, and desired target $\mathcal{X}$.
        }
        \label{fig:robot_coordinates}
\end{figure}
the robot's base frame
$\mathcal{B}$, the robot's end-effector frame $\mathcal{E}$,
and its desired target pose $\mathcal{X}$ during motion control.
$\mathcal{E}$ and $\mathcal{X}$ are each defined by a position vector
$\bm{x} = [x~y~z]^T \in \mathbb{R}^3$ and an orientation quaternion $\bm{q} = q_w + q_x i + q_y j + q_z k$, given in the base frame
$\mathcal{B}$. The superscript ${}^d$ indicates
\textit{desired} quantities for target pose tracking.
The end-effector frame $\mathcal{E}$ is located at the grasping center of the
1-DOF gripper, whose state (opening percentage) is determined by the scalar
$g \in [0, 1]$.
The vector $\bm{\theta} \in \mathbb{R}^6$ denotes the set of joint angle positions.
During motion control, a suitable controller continuously tracks
the given target $\mathcal{X}$ with the end-effector frame $\mathcal{E}$.

\subsection{Teleoperation}
\begin{figure}
        \centering
        \includegraphics[width=0.48\textwidth]{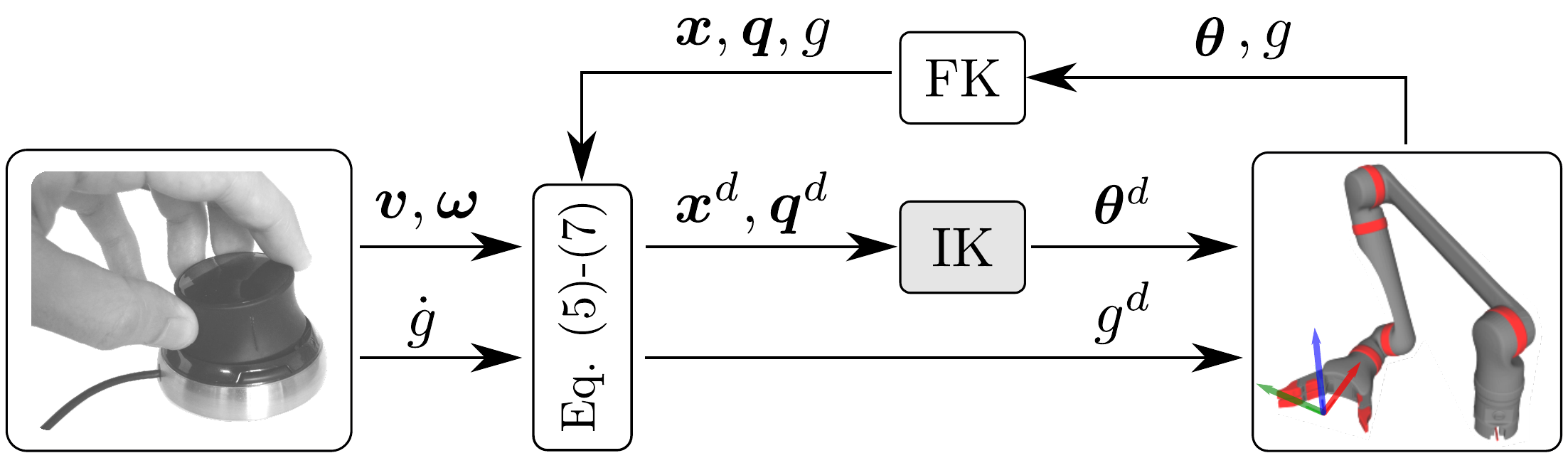}
        \caption{Teleoperation with a 6-axis joystick. We integrate linear and
        angular velocity measurements into the desired target pose $\bm{x}^d, \bm{q}^d$ for
        end-effector control. Additional buttons allow setting the opening and closing speed $\dot{g}$ for the gripper.
        }
        \label{fig:robot_teleoperation}
\end{figure}

Teleoperation must be simple and intuitive.
Our approach uses a conventional space mouse as a 6-axis joystick, but
similar low-cost devices could be used as well.
This joystick measures infinitesimal displacements from interaction with thumb
and fingers in three linear axes and three rotational
axes, and interprets those signals as a combined, six-dimensional twist command
$[\bm{v} ~ \bm{\omega}]^T = [v_x ~ v_y ~ v_z ~ \omega_x ~ \omega_y ~ \omega_z]^T$.
Starting from the robot's end-effector frame $\mathcal{E}$ in each control
cycle, we then time-integrate this twist to become the new target pose
$\mathcal{X}$ for motion tracking according to
\begin{align}
        \bm{x}^d &= \bm{x} + \bm{v} \Delta t \\
        \bm{q}^d &= \bm{q} + \dbm{q} \Delta t \quad , \qquad  \dbm{q} = \frac{1}{2} \bm{\omega} \bm{q} ~ .
\end{align}
\fig{fig:robot_teleoperation} shows the control scheme.
All quantities are given in the robot's base frame $\mathcal{B}$.
The quaternion product $\bm{\omega} \bm{q}$ uses the angular velocity given in quaternion notation
$\bm{\omega} = 0 + \omega_x i + \omega_y j + \omega_z k$.
In a similar fashion, we time-integrate the gripper's opening/closing speed,
which are triggered by pressing additional buttons on the joystick,
to obtain its target state $g^d$ with
\begin{equation}
        \label{eq:gripper_control}
        g^d = g + \dot{g} \Delta t ~ .
\end{equation}
The computation between Cartesian space and joint space is done with
suitable \textit{inverse kinematics}~(IK) and \textit{forward kinematics}~(FK) algorithms.

\subsection{Motion Tracking and Inverse Kinematics}
The robots we target support streaming joint position control interfaces, i.e.
they require $\bm{\theta}^d$ at a specific control rate.
We use a forward dynamics-based control approach of our previous
work~\cite{Scherzinger2019Inverse} for solving the IK problem during tracking
the target pose $\mathcal{X}$.
Its implementation is available open-source\footnote{https://github.com/fzi-forschungszentrum-informatik/cartesian\_controllers}.
In short, this IK solver transforms the difference between end-effector target $\mathcal{X}$ and current
pose $\mathcal{E}$ into a 6-dimensional error.
It then applies this error as a goal-directed force-torque vector
at the end-effector of a virtual model of the robot,
and closed-loop simulates the reaction motion of the system.
The result $\bm{\theta}^d$ is streamed to joint-level position controllers of the real system.
We refer the interested reader to \cite{Scherzinger2020virtual} for more details on this aspect.
\section{COMPOSING AND USING SKILLS}
\label{sec:composing_and_using_skills}
Motion skills are a central element for simplifying repetitive tasks.
This section describes the methods behind building skills from recorded teleoperation,
and details how we create new goal-oriented trajectories from these representations.

\subsection{Skill Recording and Representation}
\begin{figure*}
        \centering
        \includegraphics[width=0.85\textwidth]{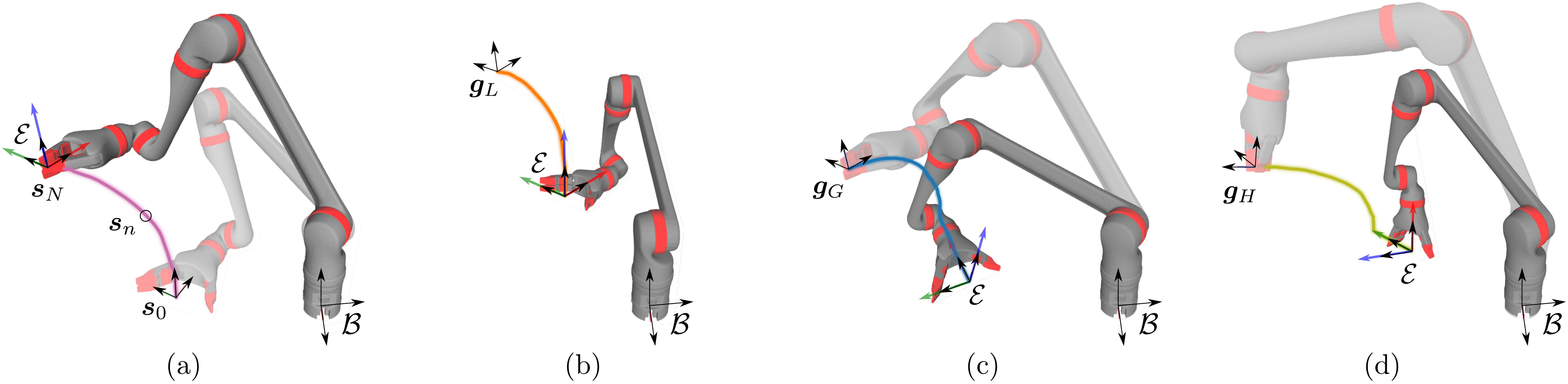}
        \caption{Skill recording (a) and generalization with three different execution types: (b)
        \textit{local} skill execution, (c) \textit{global} execution, and (d)
        \textit{hybrid} execution.
        }
        \label{fig:skill_recording_and_generalization}
\end{figure*}

We represent a specific robot state as the vector
\begin{equation}
        \label{eq:state_vector}
        \bm{s} = {}^{\mathcal{B}}[x, y, z, q_x, q_y, q_z, q_w, g]
\end{equation}
that comprises the translation and orientation of $\mathcal{E}$, and the
dimensionless gripper.
All entities are given in the robot's base frame $\mathcal{B}$, and are
computed from the joint positions $\bm{\theta}$ with forward kinematics.
Using Cartesian coordinates for the state vector - as opposed to a joint-space
representation - is important for motion generalization and shall make it easier
to anticipate the robot's motion during skill playback for new targets.
Note that we include the orientation quaternion component-wise and enforce unit length later at generalization.
Equivalent to conventional DMPs, our skills capture motion characteristics from
recorded state trajectories.
We thus first teleoperate the robot and record a sequence of states
$\{{\bm{s}_0, \bm{s}_1, \bm{s}_2, \dots , \bm{s}_N} \}$
in equidistant time steps $\Delta t$ for a certain duration $T$.
\fig{fig:skill_recording_and_generalization}(a) illustrates an example of grasping an imaginary
object and lifting it sideways in a slight arc.

We transform all recorded states with respect to the robot's
end-effector frame when recording started.
We use homogeneous matrices for transformation, and denote $\bm{T}(\bm{s})$ for
the homogeneous matrix representation of $\bm{s}$.
Due to the quaternion notation of \eq{eq:state_vector}, formulating states as
homogeneous matrices, and composing states back from the entries of a
homogeneous matrix, is unique.
With that, the states become
\begin{equation}
        \bm{s}_n \gets \bm{T}^{-1} (\bm{s}_0) \bm{T}(\bm{s}_n) ~, n = 0 \dots N ~.
\end{equation}
Note that the gripper state $g$ needs no transformation, and is simply kept
throughout all coordinate systems. The first recorded state for any skill is thus $\bm{s}_0 =
[0~0~0~0~0~0~1~g_0]^T$ with a gripper opening $g_0$.

The next step is to compute the time derivatives
$\dbm{s} = \frac{d}{dt}(\bm{s})$ and $\ddbm{s} = \frac{d}{dt}(\dbm{s})$,
for which we use
a five-step differentiator~\cite{Holoborodko2008} with
\begin{equation}
        \begin{split}
                \frac{d}{dt}(\bm{s}_n) &:= 
                \frac{5 (\bm{s}_{n+1} - \bm{s}_{n-1})}{32 \Delta t}  + \frac{4 (\bm{s}_{n+2} - \bm{s}_{n-2})}{32 \Delta t} \\
                & ~ + \frac{\bm{s}_{n+3} - \bm{s}_{n-3}}{32 \Delta t}
        \end{split}
\end{equation}

that we apply to every state $n = 0 \dots N$ in the recorded sequence.
Start and end boundaries are considered with
\begin{equation}
        \bm{s}_{(.)} =
        \begin{cases}
                \bm{s}_0 & \text{for $(.) \le 0$} \\
                \bm{s}_N & \text{for $(.) \ge N$} \\
                \bm{s}_{(.)} & \text{else} ~.
        \end{cases}
\end{equation}

We use Ijspeert's simple spring damper model~\cite{Ijspeert2013} as transformation system,
adapted to our state vector
\begin{equation}
        \ddbm{s} = \bm{D} (\bm{K} (\bm{g} - \bm{s}) - \dbm{s} ) + \bm{f} ~.
        \label{eq:skill_transformation_system}
\end{equation}
The vector $\bm{g}$ stands for the goal state of the motion and will vary with
skill \textit{type} that we explain in the next section.
The diagonal stiffness $\bm{K} \in \mathbb{R}^{8\times8}$ and damping matrix $\bm{D} \in \mathbb{R}^{8\times8}$ are
chosen once, and remain constant throughout creating and using skills.
We briefly explain how to chose them in \sect{sec:robotic_setup}.
Analog to the conventional DMP approach, we obtain $n$ unique samples for the
forcing term by solving \eq{eq:skill_transformation_system} for
$\bm{f}$ and substituting the state trajectory in
\begin{equation}
        \bm{f}_n = \ddbm{s}_n - \bm{D}( \bm{K}(\bm{g} - \bm{s}_n) - \dbm{s}_n) ~.
\end{equation}
In contrast to conventional DMPs, however, the computed sequence of forcing
terms $\{ \bm{f}_0, \bm{f}_1, \bm{f}_2, \dots
, \bm{f}_N \}$ already represents the quintessence of a skill.

\subsection{Skill Types and Generalization}

Skills are meant to be recorded at one point and replayed at another point.
We propose three different \textit{types} of skills that users can choose from to compose their
everyday tasks: \textit{local}, \textit{global}, and \textit{hybrid}.
Each type will lead to a slightly different formulation of the goal state
$\bm{g} \in \{ \bm{g}_L, \bm{g}_G, \bm{g}_H\}$ in the transformation system \eq{eq:skill_transformation_system}, and
will shape the path that the robot takes when applying the motion skill.
The next sections present $\bm{g}$ for each case.

\subsubsection{Local Skills}
This is the simplest type of generalization.
The goal state is set to the final state as seen from the end-effector frame when we started the recording:
\begin{equation}
        \bm{g}_L \gets \bm{T}^{-1} (\bm{s}_0) \bm{T}(\bm{s}_N) ~.
\end{equation}
It causes a simple playback of the recorded motion locally,
starting from the robot's momentary end-effector frame $\mathcal{E}$.
\fig{fig:skill_recording_and_generalization}(b) shows the path that the robot
would take for the example motion of
\fig{fig:skill_recording_and_generalization}(a).

\subsubsection{Global Skills}
This type of skill sets the goal state to the final state of the recording and
displays that in the current robot's end-effector frame:
\begin{equation}
        \bm{g}_G \gets \bm{T}^{-1}(\bm{s}^*) \bm{T}(\bm{s}_N) ~.
\end{equation}
It causes a generalization back towards where the recorded motion ended, seen
globally in the robot's workspace.
It starts from the robot's momentary end-effector pose, denoted here with $\bm{s}^*$.
\fig{fig:skill_recording_and_generalization}(c) shows the effect, using the previous example motion.
This skill can be used whenever the exact end position and orientation of the gripper is important.

\subsubsection{Hybrid Skills}
This type is a combination of local and global skills.
It drives back to where the recorded motion globally ended but starts the motion with the momentary end-effector orientation.
It can be used to repeatedly put objects somewhere without enforcing unnatural gripper orientations.
To achieve this, we make use of an additional frame that is adequately oriented
in the robot's end-effector frame and scale the local motion towards the global goal:
\begin{equation}
        \bm{g}_H \gets \bm{T}(\bm{g}_L) %
        \begin{bmatrix}0 \\ 0 \\ 0 \\
                \frac{\lVert \bm{x}_G \rVert}{ \lVert \bm{x}_L \rVert }
        \end{bmatrix} ~.
        \label{eq:hybrid_goal}
\end{equation}
$\bm{x}_G$ and $\bm{x}_L$ denote the translational parts of $\bm{g}_G$ and $\bm{g}_L$, respectively.
\fig{fig:skill_recording_and_generalization}(d) shows the respective path.
We describe the details of the additional frame in the next section together with how all skills
transform the created motion back to the robot's base frame $\mathcal{B}$ for
control.

\subsection{Trajectory Generation and Playback}
Creating new trajectories is done in three steps.
They all take place just before executing a skill in a new position.
First, we compute a new
sequence of states $\bm{s}_n$ by numerically integrating
\eq{eq:skill_transformation_system} with one of the goal formulations from the
previous section; then we transform this sequence back into the robot's base
frame for control; and finally assign desired timestamps for each state, and
execute the motion with the robot.

\subsubsection{Numerical integration} We integrate the transformation system with the forward Euler method.
\alg{alg:trajectory_generation} shows the scheme.
Note that only the desired duration $T$ of the skill and its type
is chosen by the user.
The other arguments are derived implicitly from the robot's current pose in its
workspace, such as the respective goal $\bm{g} \in \{ \bm{g}_L, \bm{g}_G,
\bm{g}_H\}$, and the start for motion generation $\bm{s}_0, \dbm{s}_0$.
Both $\bm{f}_n$ and $N$ are defined by the recorded skill.

\begin{algorithm}[H]
        \caption{Trajectory generation}
        \label{alg:trajectory_generation}
        \begin{algorithmic}[1]
                \Procedure{INTEGRATE}{$\bm{g}, \bm{s}_0, \dbm{s}_0, \bm{f}_n, N, T$}
                \State $\Delta t = T / N$
                \For{$n = 1 \text{ \bf{to} } N$}
                        \State $\bm{f}_{\Vert} = \frac{\bm{f} \cdot \bm{x} }{\bm{x} \cdot \bm{x}} \bm{x}$
                        \State $\bm{f}_{\perp} = \bm{f} - \bm{f}_{\Vert} $
                        \State $\bm{f}_{\Vert} = \frac{\lVert \bm{x}_G \rVert}{ \lVert \bm{x}_L \rVert } \bm{f}_{\Vert}$
                        \State $\bm{f} = \bm{f}_{\perp} + \bm{f}_{\Vert}$
                        \State $\ddbm{s}_n = \bm{D} (\bm{K} (\bm{g} - \bm{s}_{n-1}) - \dbm{s}_{n-1} ) \bm{f}_n$
                        \State $\dbm{s}_n = \dbm{s}_{n-1} + \ddbm{s}_n \Delta t$
                        \State $\bm{s}_n = \bm{s}_{n-1} + \dbm{s}_n \Delta t$
                \EndFor
                \State \textbf{return} $\{ \bm{s}_0, \bm{s}_1, \dots , \bm{s}_N\}$
        \EndProcedure
        \end{algorithmic}
\end{algorithm}
Step $4$ to $7$ rescale the translational part $\bm{f} \in \mathbb{R}^3$ of the
forcing term $\bm{f}_n \in \mathbb{R}^8$ to take the new
goal distance and the possibly changed stiffness into consideration.
This is necessary, because the new goal attractor might pull harder than
before, and might completely overrule the forcing term especially at the beginning of the
generalization.
We mitigate this effect by projecting $\bm{f}$ onto the goal direction $\bm{x} \in
\mathbb{R}^3$, scaling it, and re-adding the orthogonal component $\bm{f}_{\perp}$.

The result of \alg{alg:trajectory_generation} is a sequence of $N$ states that describe the new motion path in a skill-relative reference frame.
Note that it is important to normalize the quaternion components afterward in the individual states before succeeding operations.
\alg{alg:trajectory_generation} works dimension-wise and does not enforce this.

\subsubsection{Transformations}
This step is to display the created sequence with respect to the robot's base
$\mathcal{B}$ for the IK solver.
Both \textit{local} and \textit{global} skills generalize the motion with respect to their momentary end-effector frame.
Their motion is thus transformed with
\begin{equation}
        \label{eq:transform_to_base}
        \bm{s}_n \gets \bm{T}(\bm{s}^*) \bm{T}(\bm{s}_n) ~, n = 0 \dots N ~.
\end{equation}

In contrast, \textit{hybrid} skills generalize in a specialized reference
frame, and need an additional rotation before transforming their states back to
the robot base.
\begin{figure}
        \centering
        \includegraphics[width=0.35\textwidth]{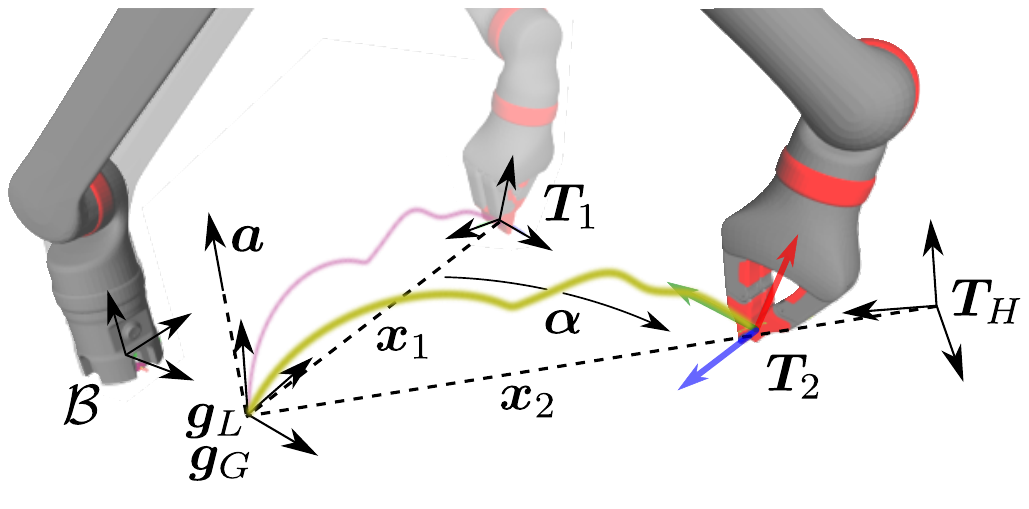}
        \caption{Hybrid skill execution with recorded motion (purple) and
        projected motion (olive) from a different start pose.
        }
        \label{fig:hybrid_skills}
\end{figure}
\fig{fig:hybrid_skills} provides a second example that we use to better illustrate the required steps.
It shows a recorded motion that we apply from a different start pose.
The two homogeneous transformation matrices
\begin{align}
        \bm{T}_1 &= \bm{T}^{-1}(\bm{g}_L) ~,\\
        \bm{T}_2 &= \bm{T}^{-1}(\bm{g}_G)
\end{align}
represent the start of the example recording and the start of the \textit{hybrid} generalization, respectively.
Note that the recorded motion's local goal $\bm{g}_L$ and the global goal
$\bm{g}_G$ always describe the same pose in the robot's workspace.
Since \textit{hybrid} skills shall drive to the global position while reproducing the motion from their local orientation,
we need to rotate the recorded motion's start into the direction of $\bm{T}_2$ with
\begin{equation}
        \bm{T}_H = \bm{T}(\bm{\alpha}, \bm{a}) \bm{T}_1 ~.
\end{equation}
\fig{fig:hybrid_skills} shows the rotation angle $\alpha$ and the rotation axis $\bm{a}$, which are
defined by the two vectors $\bm{x}_1, \bm{x}_2$, and whose components we directly obtain from
$\bm{T}_1$ and $\bm{T}_2$.
Note that the length scaling of the motion is already considered with \eq{eq:hybrid_goal}.

Finally, we extract the rotation $\bm{R} \gets \bm{T}_2 \bm{T}_H$ and
display the motion in the end-effector's coordinates.
\begin{equation}
        \bm{s}_n \gets
        \begin{bmatrix}
                \bm{R} & \bm{0} \\
                \bm{0}^T & 0
        \end{bmatrix}  \bm{T}(\bm{s}_n) ~, n = 0 \dots N ~.
\end{equation}
After this operation, the states are given in frame $\mathcal{E}$
and can be formulated in frame $\mathcal{B}$ with \eq{eq:transform_to_base}.

\subsubsection{Trajectory duration and playback}

\begin{figure}
        \centering
        \includegraphics[width=0.48\textwidth]{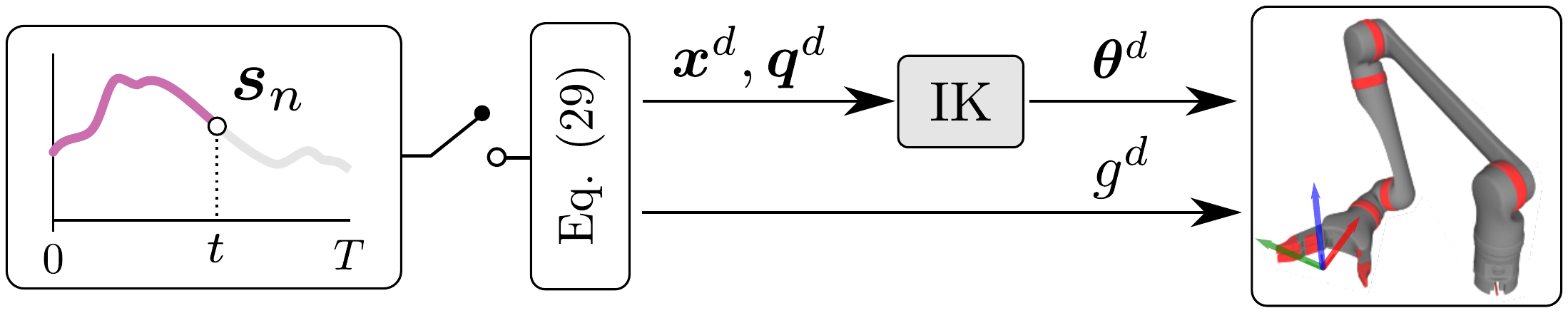}
        \caption{Skill playback on the robotic manipulator with
        sampling desired gripper poses from the created state sequence.
        Our IK solver smoothly interpolates between these discrete targets for
        the robot's joint position control.
        Note that the lower bound for $T$ is hardware specific and will vary with different manipulators.
        Values below that limit will distort and delay the end-effector motion from the desired reference.
        }
        \label{fig:skill_playback}
\end{figure}

Users can specify the duration $T$ of the skill that is fed into \alg{alg:trajectory_generation}.
This is helpful if the motion is to be reproduced slower or faster than was recorded.
Classic DMPs use a \textit{canonical system} from \eq{eq:dmp_canonical_system} for this task and
build the complete distribution of radial basis functions for approximating the forcing term $\bm{f}$ on this mechanism.
Our practical approach circumvents this complexity by simply assigning the
desired timestamp to each of the individual states of the sequence with
\begin{equation}
        \bm{s}^d(t) =
        \begin{cases}
                \bm{s}_{n} ~ \text{with} ~ n = \mathbb{N}(\frac{t}{\Delta t}) & \text{for $0 \le t < T$} \\
                \bm{s}_N & \text{for $T \le t$}
        \end{cases} ~.
\end{equation}
The notation $\mathbb{N}(.)$ denotes a truncation and conversion into a natural
number $n \in \mathbb{N}$.
Each of the individual states $\bm{s}^d$ holds the desired robot control
commands according to \eq{eq:state_vector}, and is passed to our IK solver as depicted in \fig{fig:skill_playback}.

\section{EXPERIMENTS AND RESULTS}       %
\label{sec:experiments_and_results}
This section demonstrates the feasibility of the proposed methods
with applying \textit{local}, \textit{hybrid}, and \textit{global} skills to example use cases.
For each, we record a skill by teleoperating the real robot through the motion once
and apply the skill from various starting poses.
We then plot and discuss the end-effector paths that the robot takes.
The executions of all motions with the robot are shown in the accompanying video.
In addition to this evaluation, we also give practical hints on how to
mentally map everyday tasks to these three skill types.

\subsection{Robotic Setup}
\label{sec:robotic_setup}
We use the \textit{Accrea Aria} robot for the experiments.
It has an integrated gripper with one degree of freedom and three passive compliant fingers.
\fig{fig:experiment_local_skills}(a) shows the arm in an upright position.
It is compatible with ROS-control~\cite{Chitta2017}, and provides the necessary
streaming interface for joint positions\footnote{https://github.com/stefanscherzinger/aria-ros/tree/devel}.
We use a \textit{3DConnexion} spacemouse as joystick for teleoperating the robot.
According to ~\cite{Ijspeert2013}, \eq{eq:skill_transformation_system} 
is critically damped for $\bm{D} = 4 \bm{K}$,
and we set slightly over-critical damping with $\bm{K} = 0.55 \cdot
\bm{I}^{8\times8}$ and $\bm{D} = 3.5 \cdot \bm{I}^{8\times8}$.

\subsection{Local Skill Evaluation}

\begin{figure}
        \centering
        \begin{subfigure}[b]{0.15\textwidth}
                \includegraphics[width=1.0\textwidth]{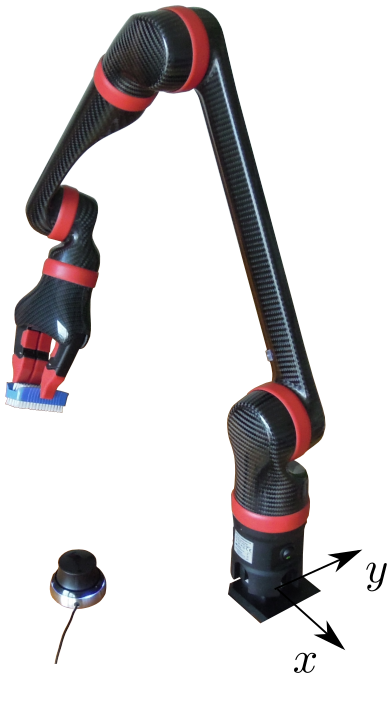}
                \caption{}
        \end{subfigure}%
        \begin{subfigure}[b]{0.25\textwidth}
                \includegraphics[width=1.0\textwidth]{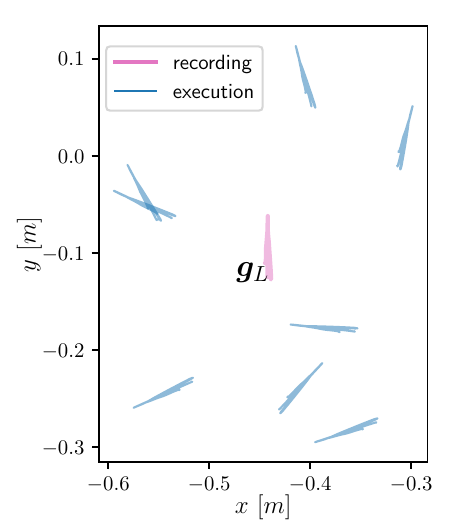}
                \caption{}
        \end{subfigure}
        \caption{(a): The robotic manipulator from our experiments with
        the joystick for teleoperation. (b) Brush movements on a flat surface,
        applied with a \textit{local} skill at different positions and orientations.
        }
        \label{fig:experiment_local_skills}
\end{figure}

Local skills simply replay the recorded motion in the robot's current end-effector frame.
This is useful for skills that can be formulated entirely central to some spot, such as wiping, brushing, or scratching.
\fig{fig:experiment_local_skills} shows a recorded brushing motion on a flat
surface and various skill replays at slightly different spots.
The curves show that the robot's end-effector replays the recorded pattern as expected.

\subsection{Hybrid Skill Evaluation}
\begin{figure}
        \centering
        \begin{subfigure}[b]{0.40\textwidth}
                \includegraphics[width=1.0\textwidth]{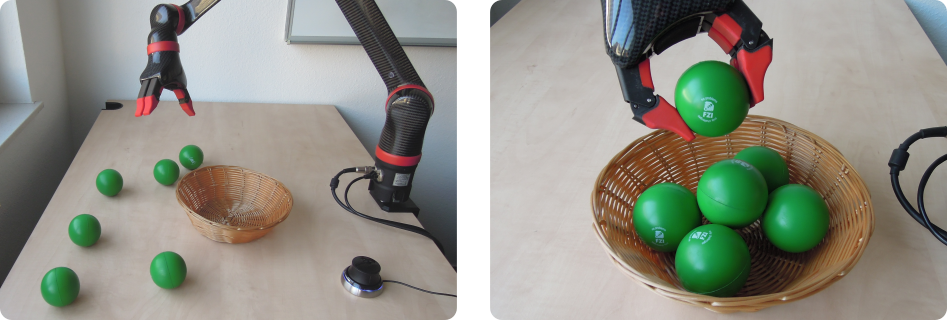}
                \caption{}
        \end{subfigure}
        \begin{subfigure}[b]{0.40\textwidth}
                \includegraphics[width=1.0\textwidth]{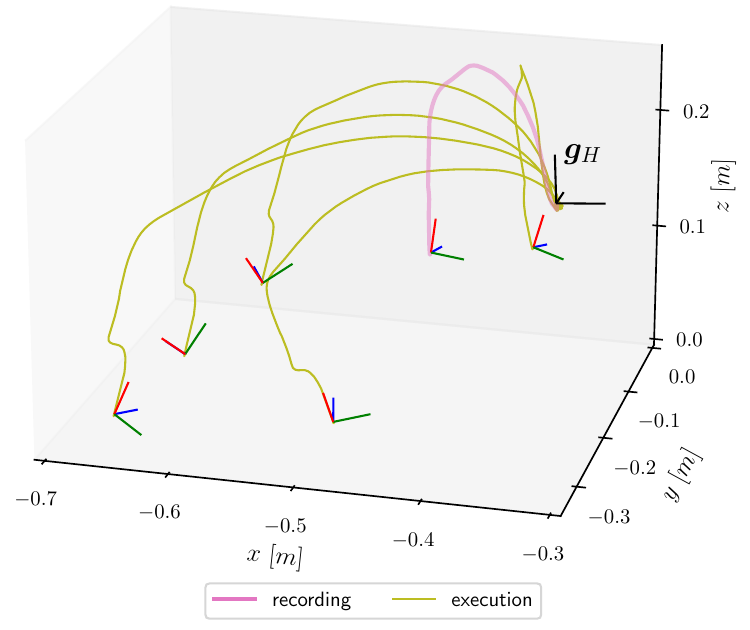}
                \caption{}
        \end{subfigure}
        \caption{(a) Collecting objects in a basket.
        (b) Recorded paths of the robot's end-effector with \textit{hybrid}
        skill execution.
        A single recording is generalized from different starting poses. 
        }
        \label{fig:experiment_hybrid_skills}
\end{figure}

This experiment evaluates our methods for goal-directed motions from different starting poses.
\fig{fig:experiment_hybrid_skills}(a) shows the setting.
The robot is clamped to the table and
the task is to collect softballs in a basket.
If successful, all skill executions must terminate where the recording ended,
independent from their start orientation, and the traveled path should
reflect the characteristics of the recording.
We conducted the experiment as follows:
First, we teleoperated the robot once to record the skill.
The recording started right before grasping the softball and ended after dropping it above the basket.
We then teleoperated the robot manually to the remaining softballs and
triggered the previously recorded skill at each spot with an open gripper.
\fig{fig:experiment_hybrid_skills}(b) shows the individual skill executions.
The coordinate systems at the starts illustrate the gripper's orientation before grasping.
All curves terminate in the hybrid goal $\bm{g}_H$, as was intended,
and the initial motion pattern is successfully scaled and reproduced with their orientation.

\subsection{Global Skill Evaluation}

\begin{figure}
        \centering
        \begin{subfigure}[b]{0.40\textwidth}
                \includegraphics[width=1.0\textwidth]{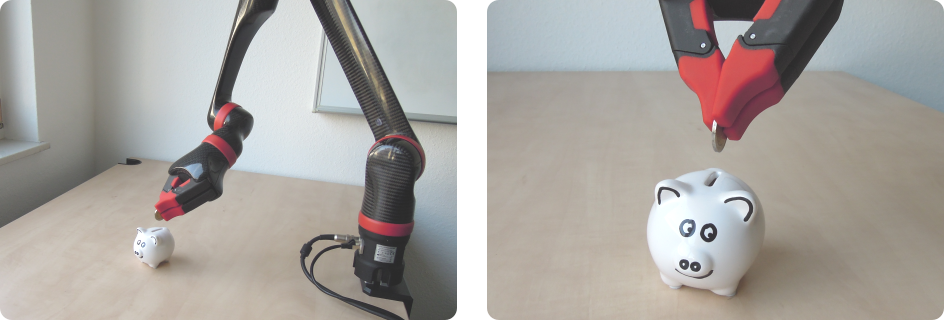}
                \caption{}
        \end{subfigure}

        \begin{subfigure}[b]{0.41\textwidth}
                \includegraphics[width=1.0\textwidth]{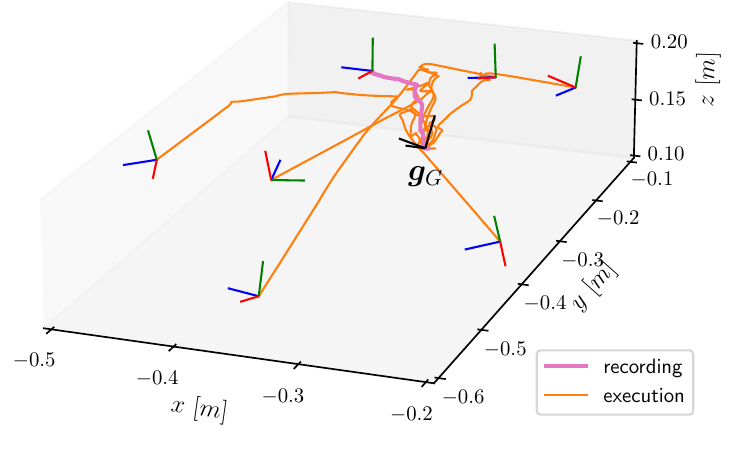}
                \caption{}
        \end{subfigure}
        \caption{(a) Feeding coins into a piggy bank from different starts. The
        coin's final orientation is essential for success.
        (b) Recorded paths of the \textit{global} skill execution on the robot.
        }
        \label{fig:experiment_global_skills}
\end{figure}

This last experiment shows the generalization capabilities to exact goal poses.
We feed a piggy bank as shown in \fig{fig:experiment_global_skills}(a), where
the coin must be carefully aligned with the opening to succeed.
While in the previous experiment the gripper mostly kept its local orientation during
motion execution, this skill needs to generalize across
the different quaternion dimensions to finally converge to the absolute goal pose.
\fig{fig:experiment_global_skills}(b) shows the effectiveness of the presented methods.
Note that the recorded motion was comparatively short and described a slight arc above the piggy.
The skills, nevertheless, worked even from starts
below this level and from significantly greater distances.

\subsection{Skill Learning Performance}
\begin{figure}
        \centering
        \includegraphics[width=0.48\textwidth]{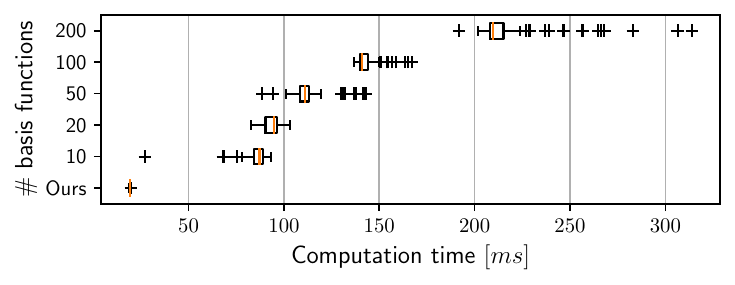}
        \caption{
                Cost of learning the recorded reference motion from~\fig{fig:experiment_global_skills}.
                We compared our approach against Ginesi et al~\cite{Ginesi2021}
                with a varying number of Gaussian basis functions.
                Each boxplot shows 100 runs, computed on an
                Intel\textsuperscript{\textregistered}
                Core\texttrademark~i7-4900MQ.
        }
        \label{fig:skill_learning_performance}
\end{figure}

\begin{figure}
        \centering
        \includegraphics[width=0.49\textwidth]{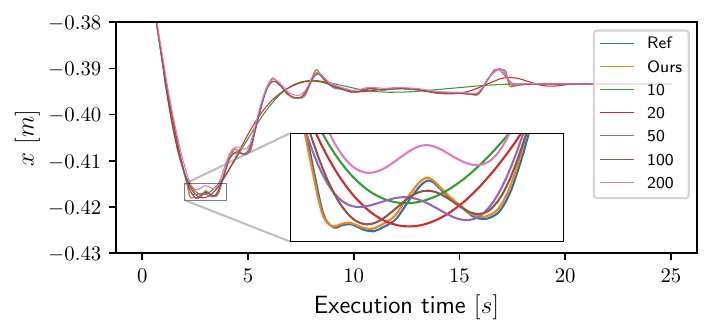}
        \caption{
                Accuracy of imitating the recorded reference motion
                from~\fig{fig:experiment_global_skills} for a selected
                dimension.
        }
        \label{fig:skill_accuracy}
\end{figure}

\begin{table}[htbp]
\centering
\caption{Skill learning accuracy}
\label{tab:skill_accuracy}

        \begin{tabular}{rcccccc}
\toprule
                \# Gaussians & 10 & 20 & 50 & 100 & 200 & Ours \\
\midrule
                rmse [mm] & 1.46 & 1.3 & 0.317 & 0.224 & 0.705 & \textbf{0.109} \\
                max [mm] & 5.3 & 4.69 & 1.27 & 1.42 & 3.05 & \textbf{0.406} \\
                std-dev. [mm] & 1.1 & 1.04 & 0.259 & 0.19 & 0.533 & \textbf{0.0911} \\

\bottomrule
\end{tabular}
\end{table}

In this experiment, we compared our approach against Ginesi et al.~\cite{Ginesi2021}, who use radial basis functions for approximating the forcing term.
Their implementation\footnote{https://github.com/mginesi/dmp\_pp} is open-source, and lets us choose Gaussians according to~\eq{eq:dmp_forcing_term}.
\fig{fig:skill_learning_performance} shows the computational cost for a selected number of basis functions.
Although skill learning is relatively fast in all cases, our approach is roughly one order of magnitude faster than the $100$-model.
Note that in the conventional approach, the number of basis functions directly correlates with the accuracy, with which the learned skills approximate the recorded motion.
\fig{fig:skill_accuracy} shows this influence on one of the path's dimensions.
Note how the $200$-model worsens the path approximation in comparison to the $100$-model.
\tabl{tab:skill_accuracy} shows relevant statistics for these curves, such as the root mean
squared error (rmse), the maximal, absolute path deviation (max), and the standard
deviation (std-dev.).
In conclusion, our approach provides the fastest skill learning performance
while best capturing detailed motion characteristics in the recordings.

\subsection{Discussion}
Within our proposed framework, all goals are implicitly given through the last state of the motion recording.
We think that this is a suitable trade-off between flexibility and ease of use, since it
avoids that users need to specify Cartesian poses in the robot's
workspace, and instead record what they want via teleoperation.
Saving the forcing terms directly as the quintessence of motion skills means higher disk space in comparison to
saving the parameters of the basis functions in the classic DMP approach.
We think that this drawback is justified, however, through not requiring to learn
those parameters online so that teleoperating and creating skills on-the-fly is
as fast as possible.
Note that the presented methods do not aim for assistive autonomy,
and there is no plausibility check whether the robot can execute a skill in the specified
duration, nor whether a skill's motion might get distorted through singular
joint configurations.
Instead, our approach is to give users cognitive responsibility and
freedom over recording and deploying skills as they please, including to learn from trial and error.

Here are some examples to help how to think everyday tasks in our three categories:
\begin{itemize}
        \item \textit{Local skills}: Spooning food, opening doors and drawers, shaking a pack of juice, pouring, scratching an itchy spot
        \item \textit{Hybrid skills}: Throwing things in a trash bin, putting ingredients in a pot, removing captured stones from a \textit{Go} board
        \item \textit{Global skills}: Grasping and passing objects, bringing food to one's mouth
\end{itemize}

Also, note that \textit{hybrid} and \textit{global} skills are mostly short-lived and
make sense to be recorded immediately before executing a repetitive task.
They could, however, be part of a permanent skill repertoire if the robot is mounted to a
wheelchair, for instance, and the goals do not change with respect to the
robot's base.

\section{CONCLUSIONS}
\label{sec:conclusions}

This paper presented a simplified DMPs-based approach for recording and
playing-back skills with teleoperated, assistive robot arms.
We proposed three skill types that can serve as a modular basis for simplifying repetitive patterns of everyday tasks.
The skills take advantage of start and end states during recording and
circumvent the difficulty of specifying goal attractors manually.
By using the DMPs' forcing term directly on the transformation system,
we cut the complexity of parameterization with basis functions, data fitting, and the phase variable and instead realized time
scaling with a simple reassigning of timestamps to fixed-step trajectories.
This assures a minimal computational runtime cost during
skill creation whithout sacrificing any accuracy in the recordings.
The trajectories are then executed open-loop by an interpolating Cartesian
controller that is robot-agnostic for 6-axes manipulators.
The presented methods target assistive robot arms without requiring sensors for
perception and are suitable whenever users need a simple, tool-like mechanism to
compose and use skills during teleoperation.

\renewcommand*{\bibfont}{\small}
\printbibliography

\end{document}